\documentclass{article}
\usepackage{spconf,amsmath,graphicx}

\usepackage{booktabs} 
\usepackage{multirow}
\usepackage[font=small]{caption}
\usepackage{url}
\usepackage{amsfonts}

\def\etal{\emph{et al}.}
\def\eg{\emph{e.g}.}
\def\ie{\emph{i.e}.}

\title{ITERATIVE SELF KNOWLEDGE DISTILLATION --- FROM POTHOLE CLASSIFICATION TO FINE-GRAINED AND COVID RECOGNITION}
\name{Kuan-Chuan Peng}
\address{Mitsubishi Electric Research Laboratories (MERL), Cambridge, MA, USA\\
{\tt\small kpeng@merl.com}
}
\begin{document}
%
\maketitle
\begin{abstract}
Pothole classification has become an important task for road inspection vehicles to save drivers from potential car accidents and repair bills. Given the limited computational power and fixed number of training epochs, we propose iterative self knowledge distillation (ISKD) to train lightweight pothole classifiers. Designed to improve both the teacher and student models over time in knowledge distillation, ISKD outperforms the state-of-the-art self knowledge distillation method on three pothole classification datasets across four lightweight network architectures, which supports that self knowledge distillation should be done iteratively instead of just once. The accuracy relation between the teacher and student models shows that the student model can still benefit from a moderately trained teacher model. Implying that better teacher models generally produce better student models, our results justify the design of ISKD. In addition to pothole classification, we also demonstrate the efficacy of ISKD on six additional datasets associated with generic classification, fine-grained classification, and medical imaging application, which supports that ISKD can serve as a general-purpose performance booster without the need of a given teacher model and extra trainable parameters.
\end{abstract}

\section{Introduction}

Detecting potholes is essential for the municipalities and road authorities to repair defective roads. The vehicle repair bills related to pothole damage have cost U.S. drivers \$3 billion annually on average~\cite{AAA_PFS}. Due to the cost constraint of the edge devices which the pothole classifiers run on, the edge devices installed on the road inspection vehicles may only have limited computational power (\eg, no GPU). In such scenarios, lightweight models are needed if real-time inference speed is required. Motivated by this application and deployment time constraint of pothole classifiers, we focus on the following problem: \textit{Given a fixed number of training epochs and a lightweight model to be trained, what can practitioners do to improve the pothole classification accuracy}?

Given the problem, we explore \textit{self} knowledge distillation (KD)~\cite{Yuan_CVPR20} to tackle pothole classification. By \textit{self} KD, we refer to the KD methods which need no teacher model in advance and introduce no extra trainable parameters. Showing that KD is actualy learned label smoothing regularization, Yuan~\etal~\cite{Yuan_CVPR20} propose Tf-KD$_{self}$, the teacher-free KD by using the pre-trained student model itself as the teacher model. Inspired by~\cite{Yuan_CVPR20} and the assumption that better teacher models result in better student models, we propose \textit{iterative self knowledge distillation (ISKD)}, which iteratively performs self KD by using the pre-trained student model as the teacher model.

Most KD methods~\cite{Passalis_CVPR20,Zhao_CVPR20,Xu_ECCV20,Peng_ICCV19,Jin_ICCV19} typically require that the teacher model is available in advance, which is not always true. Even if there are KD methods which need no teacher model~\cite{Yuan_CVPR20,Guo_CVPR20,Yun_CVPR20}, these methods typically do not experiment on lightweight models or perform KD multiple times. Utilizing KD iteratively and requiring no teacher model, our proposed ISKD shows its efficacy on lightweight models for pothole classification, generic classification, fine-grained classification, and medical imaging application. Although there exist methods utilizing iterative KD~\cite{Koutini_DCASE18,Furlanello_ICML18,Zhang_MM20}, they typically require additional constraints and are validated on only few datasets. For example, Koutini's method~\cite{Koutini_DCASE18} requires training multiple models and selecting the best trained model for each class to predict pseudo labels for sound event detection. In contrast, our proposed ISKD only needs to train one model at once with no need to select models and predict pseudo labels. In~\cite{Furlanello_ICML18,Zhang_MM20}, their teacher model is trained until convergence for each KD iteration, and their methods are validated on only few datasets. In addition, the accuracy gain of~\cite{Furlanello_ICML18} comes from the ensemble of all the students in the history, which needs more deployment space at testing time. In contrast, we show ISKD's efficacy on a wide variety of datasets even when the teacher model is only moderately trained, and ISKD does not rely on the ensemble, which is more practical for embedded devices.

To the best of our knowledge, we are the first in pothole classification to use iterative self knowledge distillation when training lightweight neural networks under limited training epochs. We make the following contributions:
\\\textbf{(1)} We propose iterative self knowledge distillation (ISKD), which outperforms the state-of-the-art self KD method Tf-KD$_{self}$~\cite{Yuan_CVPR20} on the road damage dataset~\cite{RDD}, the Nienaber potholes simplex~\cite{simplex_dataset} and complex~\cite{complex_dataset} datasets, CIFAR-10~\cite{cifar}, CIFAR-100~\cite{cifar}, Oxford 102 Flower~\cite{oxford102}, Oxford-IIIT Pet Dataset~\cite{oxford37}, Caltech-UCSD Birds 200~\cite{cub200}, and COVID-19 Radiography~\cite{covid} datasets, which supports the wide applicability of ISKD from pothole classification to generic, fine-grained, and medical imaging classification.
\\\textbf{(2)} We provide more evidence showing that even using a teacher model with accuracy lower than the baseline accuracy from a classifier trained with a larger number of epochs, the student model can still possibly outperform the baseline.
\\\textbf{(3)} ISKD can outperform the baseline under a wide range of weight balancing the objectives of ISKD, which supports that ISKD is flexible with respect to parameter selection.

\section{Iterative self knowledge distillation}
\label{subsec:method_ISKD}

\begin{figure}
\centering
\includegraphics[width=.745\linewidth]{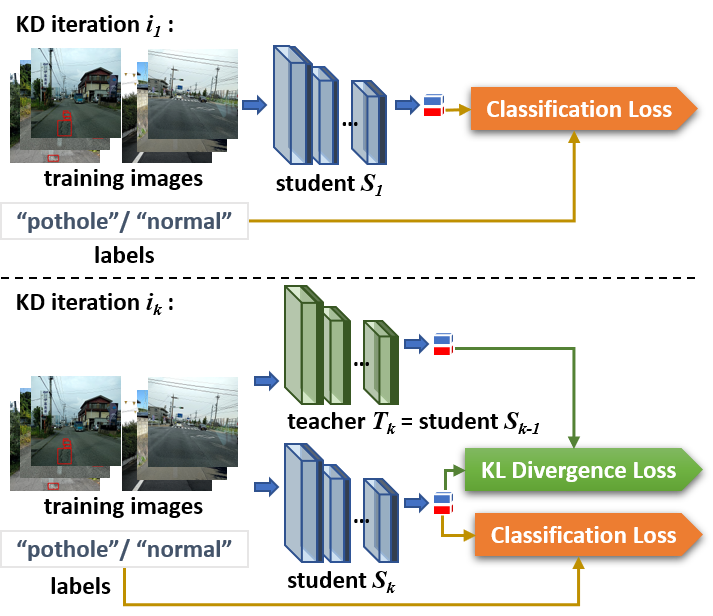}
\vspace{-.5em}
\caption{Our proposed iterative self knowledge distillation.}
\label{ISKD}
\vspace{-1em}
\end{figure}

Inspired by Yuan \etal~\cite{Yuan_CVPR20}, we propose iterative self knowledge distillation (ISKD) such that both teacher and student models can improve over time. We illustrate ISKD in Fig.~\ref{ISKD}, where we denote the teacher/student model in the $k$-th KD iteration $i_k$ as $T_k$/$S_k$. During $i_1$, since $T_1$ is not given in advance, we train $S_1$ using the softmax cross-entropy loss $L_c$ as the classification loss. During $i_k\left ( k>1,k\in \mathbb{N} \right )$, we use the trained student model in $i_{k-1}$ (\ie, $S_{k-1}$) as $T_k$, and train $S_k$ with both $L_c$ and the Kullback-Leibler (KL) divergence loss. Specifically, the total loss function to train $S_k$ during $i_k$ can be written as $L_{KD}=\left ( 1-\alpha \right )L_c + \alpha KLD\left ( \mathbf{z}, \mathbf{z}^t \right )$, where $KLD$ is the KL divergence, $\mathbf{z}^t$/$\mathbf{z}$ is the output probability distribution of $T_k$/$S_k$, and $\alpha$ is the weight of $KLD$.

During $i_k$, we freeze the parameters of $T_k$ and only train $S_k$. We pre-train $S_k$ from ImageNet~\cite{imagenet}, not from $S_{k-1}$ because we hope to decrease the chance that $S_k$ is trapped from the possibly local optimum associated with $S_{k-1}$. ISKD stops at $i_k$ if $S_k$ shows no obvious accuracy gain over $S_{k-1}$. Since Yuan \etal~\cite{Yuan_CVPR20} show that to benefit the student model, the teacher model need not outperform the student model, we directly use the previously trained student model as the current teacher model, waiving the typical KD requirement that the teacher model is needed in advance. We expect that using ISKD improves both $T_k$ and $S_k$ when $k$ increases under the assumption that using better teacher models in KD generally results in better student models.

\section{Experimental setup}

We experiment on road damage dataset (termed as RDD)~\cite{RDD}, Nienaber potholes simplex (termed as simplex)~\cite{simplex_dataset} and complex (termed as complex)~\cite{complex_dataset} datasets, CIFAR-10~\cite{cifar}, CIFAR-100~\cite{cifar}, Oxford 102 Flower dataset (termed as Oxford-102)~\cite{oxford102}, Oxford-IIIT Pet dataset (termed as Oxford-37)~\cite{oxford37}, Caltech-UCSD Birds 200 dataset (termed as CUB-200)~\cite{cub200}, and COVID-19 Radiography dataset (termed as COVID)~\cite{covid}. We choose these datasets to cover a diverse range of task domains from pothole, generic, fine-grained to  medical imaging classification. The RDD, simplex, and complex datasets provide annotations of whether each image contains any pothole or not. The Oxford-102, Oxford-37, and CUB-200 datasets provide the images and labels of 102, 37, and 200 different species of flowers, cats and dogs, and birds, respectively. The COVID dataset includes 4 different types of chest x-rays: normal, COVID, lung opacity, and viral pneumonia. For all the datasets except COVID, we use the official training/testing split of each dataset. Since the COVID dataset does not provide the official training/testing split, we randomly generate the split using the ratio of 7:3.

We use the official PyTorch~\cite{pytorch} implementation of ResNet-18~\cite{resnet}, SqueezeNet v1.1~\cite{squeezenet}, and ShuffleNet v2 x0.5 \& x1.0~\cite{shufflenetv2}, modify their last layers such that the number of output nodes of the last layer equals the number of classes, and pre-train them from the ImageNet~\cite{imagenet}. The four network architectures are selected based on the following criteria: (1) For the ease of reproducibility, they are officially supported by PyTorch~\cite{pytorch}, which provides their weights pre-trained from the ImageNet~\cite{imagenet}. (2) Considering typically limited computational power on edge devices, we limit the number of network parameters to be fewer than 12M.

We first experiment on the four network architectures mentioned previously using the RDD~\cite{RDD}, simplex~\cite{simplex_dataset}, and complex~\cite{complex_dataset} datasets. For each KD iteration, we use the same network architecture for both teacher and student models. We compare ISKD with the following two baselines with the same network architecture, total number of training epochs, and learning schedule: (1) training a classifier using $L_c$ without KD (termed as the large-epoch baseline), and (2) Tf-KD$_{self}$~\cite{Yuan_CVPR20}, which only performs KD once without multiple KD iterations. For the extended study involving the other six datasets irrelevant to potholes, we use the ResNet-18~\cite{resnet} and ShuffleNet v2 x1.0~\cite{shufflenetv2} as the network architectures of ISKD. We conduct the extended study in the same way as the pothole classification task mentioned previously using the same two baselines.

\begin{table*}[t]
\centering
\footnotesize
\begin{tabular}{@{\hspace{0em}}c@{\hspace{0.5em}}c@{\hspace{0.5em}}c@{\hspace{0.2em}}c@{\hspace{0.8em}}c@{\hspace{0.8em}}c@{\hspace{0.8em}}c@{\hspace{0.8em}}c@{\hspace{0.8em}}c@{\hspace{0.8em}}c@{\hspace{0.8em}}c@{\hspace{0em}}}
\toprule
\multirow{2}{*}{dataset} &experiment ID &$E_1$ &$E_2$ &$E_3$ &$E_4$ &$E_5$ &$E_6$ &$E_7$ &$E_8$ &$E_9$\\
&model $\backslash$ KD iteration &$i_1$ (no KD) &$i_2$ &$i_3$ &$i_4$ &$i_5$ &$i_6$ &$i_1\sim i_6$ &$i_1$ (large-epoch) &$i_1$ + Tf-KD$_{self}$~\cite{Yuan_CVPR20}\\
\midrule
\multirow{4}{*}{RDD~\cite{RDD}} &ResNet-18~\cite{resnet} &91.54$_{50}$ &92.71$_{50}$ &92.99$_{50}$ &93.04$_{50}$ &\textbf{93.08$_{50}$} &n/a &\textbf{93.08$_{250}$} &92.24$_{250}$ &92.34$_{250}$\\
&SqueezeNet v1.1~\cite{squeezenet} &89.67$_{50}$ &89.91$_{50}$ &90.28$_{50}$ &90.51$_{50}$ &\textbf{90.70$_{50}$} &\textbf{90.70$_{50}$} &\textbf{90.70$_{300}$} &90.47$_{300}$ &90.28$_{300}$\\
&ShuffleNet v2 x0.5~\cite{shufflenetv2} &90.05$_{50}$ &90.14$_{50}$ &90.98$_{50}$ &91.40$_{50}$ &91.40$_{50}$ &n/a &91.40$_{250}$ &\textbf{92.66$_{250}$} &91.22$_{250}$\\
&ShuffleNet v2 x1.0~\cite{shufflenetv2} &92.01$_{50}$ &92.15$_{50}$ &92.66$_{50}$ &\textbf{93.22$_{50}$} &\textbf{93.22$_{50}$} &n/a &\textbf{93.22$_{250}$} &93.13$_{250}$ &93.13$_{250}$\\
\midrule
\multirow{4}{*}{simplex~\cite{simplex_dataset}} &ResNet-18~\cite{resnet} &81.85$_{50}$ &90.46$_{50}$ &93.69$_{50}$ &96.92$_{50}$ &98.62$_{50}$ &\textbf{99.08$_{50}$} &\textbf{99.08$_{300}$} &83.38$_{300}$ &92.00$_{300}$\\
&SqueezeNet v1.1~\cite{squeezenet} &81.85$_{50}$ &86.77$_{50}$ &87.69$_{50}$ &\textbf{88.15$_{50}$} &\textbf{88.15$_{50}$} &n/a &\textbf{88.15$_{250}$} &84.62$_{250}$ &87.69$_{250}$\\
&ShuffleNet v2 x0.5~\cite{shufflenetv2} &90.00$_{50}$ &95.69$_{50}$ &99.38$_{50}$ &\textbf{100.00$_{50}$} &n/a &n/a &\textbf{100.00$_{200}$} &92.46$_{200}$ &97.54$_{200}$\\
&ShuffleNet v2 x1.0~\cite{shufflenetv2} &90.31$_{50}$ &96.31$_{50}$ &98.77$_{50}$ &\textbf{100.00$_{50}$} &n/a &n/a &\textbf{100.00$_{200}$} &93.38$_{200}$ &97.54$_{200}$\\
\midrule
\multirow{4}{*}{complex~\cite{complex_dataset}} &ResNet-18~\cite{resnet} &61.92$_{50}$ &74.14$_{50}$ &77.65$_{50}$ &79.80$_{50}$ &83.77$_{50}$ &\textbf{84.93$_{50}$} &\textbf{84.93$_{300}$} &62.58$_{300}$ &82.95$_{300}$\\
&SqueezeNet v1.1~\cite{squeezenet} &59.27$_{50}$ &70.70$_{50}$ &\textbf{76.16$_{50}$} &\textbf{76.16$_{50}$} &n/a &n/a &\textbf{76.16$_{200}$} &62.25$_{200}$ &71.03$_{200}$\\
&ShuffleNet v2 x0.5~\cite{shufflenetv2} &56.79$_{50}$ &78.97$_{50}$ &88.58$_{50}$ &\textbf{88.91$_{50}$} &n/a &n/a &\textbf{88.91$_{200}$} &65.89$_{200}$ &79.80$_{200}$\\
&ShuffleNet v2 x1.0~\cite{shufflenetv2} &60.43$_{50}$ &78.31$_{50}$ &\textbf{86.59$_{50}$} &86.42$_{50}$ &n/a &n/a &86.42$_{200}$ &71.85$_{200}$ &82.45$_{200}$\\
\midrule
\multirow{2}{*}{CIFAR-10~\cite{cifar10}} &ResNet-18~\cite{resnet} &90.75$_{50}$	&91.81$_{50}$	&92.05$_{50}$	&\textbf{92.07$_{50}$}	&n/a	&n/a    &\textbf{92.07$_{200}$}	&91.53$_{200}$	&92.01$_{200}$\\
&ShuffleNet v2 x1.0~\cite{shufflenetv2} &82.63$_{50}$	&85.16$_{50}$	&88.45$_{50}$	&90.30$_{50}$	&91.03$_{50}$	&\textbf{91.58$_{50}$}	&\textbf{91.58$_{300}$}  &90.68$_{300}$	&85.55$_{300}$\\
\midrule
\multirow{2}{*}{CIFAR-100~\cite{cifar10}} &ResNet-18~\cite{resnet} &80.15$_{50}$	&81.05$_{50}$	&81.64$_{50}$	&82.17$_{50}$	&82.30$_{50}$	&\textbf{82.67$_{50}$}  &\textbf{82.67$_{300}$}	&81.34$_{300}$	&81.62$_{300}$\\
&ShuffleNet v2 x1.0~\cite{shufflenetv2} &58.95$_{50}$	&65.60$_{50}$	&72.16$_{50}$	&75.59$_{50}$	&77.27$_{50}$	&\textbf{77.85$_{50}$}  &\textbf{77.85$_{300}$}	&77.61$_{300}$	&65.78$_{300}$\\
\midrule
\multirow{2}{*}{Oxford-102~\cite{oxford102}} &ResNet-18~\cite{resnet} &96.58$_{50}$	&97.31$_{50}$	&97.68$_{50}$	&\textbf{97.80$_{50}$}	&n/a	&n/a    &\textbf{97.80$_{200}$}	&96.94$_{200}$	&97.43$_{200}$\\
&ShuffleNet v2 x1.0~\cite{shufflenetv2} &94.74$_{50}$	&97.19$_{50}$	&98.17$_{50}$	&\textbf{98.41$_{50}$}	&n/a	&n/a	&\textbf{98.41$_{200}$}  &\textbf{98.41$_{200}$} 	&97.19$_{200}$\\
\midrule
\multirow{2}{*}{Oxford-37~\cite{oxford37}} &ResNet-18~\cite{resnet} &90.57$_{50}$	&90.98$_{50}$	&91.33$_{50}$	&\textbf{91.80$_{50}$}	&91.58$_{50}$  &n/a    &\textbf{91.80$_{200}$}	&91.20$_{200}$	&91.31$_{200}$\\
&ShuffleNet v2 x1.0~\cite{shufflenetv2} &79.69$_{50}$	&84.30$_{50}$	&85.96$_{50}$	&\textbf{86.59$_{50}$}	&\textbf{86.59$_{50}$}	&n/a    &\textbf{86.59$_{250}$}	&86.10$_{250}$	&84.46$_{250}$\\
\midrule
\multirow{2}{*}{CUB-200~\cite{cub200}} &ResNet-18~\cite{resnet} &42.07$_{50}$	&46.49$_{50}$	&48.66$_{50}$	&\textbf{49.82$_{50}$}	&49.49$_{50}$	&n/a    &\textbf{49.82$_{200}$}	&46.03$_{200}$	&47.58$_{200}$\\
&ShuffleNet v2 x1.0~\cite{shufflenetv2} &41.54$_{50}$	&45.40$_{50}$	&48.53$_{50}$	&49.69$_{50}$	&\textbf{50.28$_{50}$}	&49.95$_{50}$  &\textbf{50.28$_{250}$}	&48.99$_{250}$	&46.95$_{250}$\\
\midrule
\multirow{2}{*}{COVID~\cite{covid}} &ResNet-18~\cite{resnet} &94.11$_{50}$	&94.68$_{50}$	&94.80$_{50}$	&\textbf{95.01$_{50}$}	&n/a	&n/a	&\textbf{95.01$_{200}$}  &94.79$_{200}$	&94.88$_{200}$\\
&ShuffleNet v2 x1.0~\cite{shufflenetv2} &90.76$_{50}$	&92.43$_{50}$	&93.07$_{50}$	&93.81$_{50}$	&\textbf{93.98$_{50}$}	&n/a    &\textbf{93.98$_{250}$}	&92.72$_{250}$	&93.10$_{250}$\\
\bottomrule
\end{tabular}
\vspace{-.5em}
\caption{Comparing the classification accuracy (\%) of the iterative self knowledge distillation (KD) method versus the baselines. The numbers are in the format of [accuracy]$_{[e_s]}$, where $e_s$ is the number of epochs which the student model is trained for.
}
\label{ISKD_result}
\vspace{-.8em}
\end{table*}

For all the experiments, the training images are resized to 224$\times$224, and the model is pre-trained from ImageNet~\cite{imagenet} and fine-tuned with the training data of each dataset. We use the momentum 0.9, weight decay 5e-4, batch size 128, and the SGD optimizer to train the student model for 50 epochs for each KD iteration, and the learning rate is fixed within each KD iteration but model-specific during training. For ResNet-18~\cite{resnet} and SqueezeNet v1.1~\cite{squeezenet}, we use the initial learning rate 0.001, but for ShuffleNet v2 x0.5 \& x1.0~\cite{shufflenetv2}, we use the initial learning rate 0.1. Following Tf-KD$_{self}$~\cite{Yuan_CVPR20}, we obtain the $\alpha$ values by grid search on the validation data sampled from the training set when experimenting on the RDD dataset~\cite{RDD}. Once we determine the $\alpha$ values from the RDD dataset, we fix the $\alpha$ values and use the same set of $\alpha$ values when experimenting on other datasets (\ie, the $\alpha$ values are not tuned for most of the datasets except the RDD dataset). We purposely do so to test whether the $\alpha$ values searched from one dataset can be transferable and directly applied to other datasets. For all the other parameters, we use the default PyTorch~\cite{pytorch} setting unless otherwise specified.

\begin{table}
\centering
\small
\begin{tabular}{@{}r@{\hspace{1em}}c@{\hspace{1.5em}}l@{}}
\toprule
dataset$\backslash$method &ISKD &prior work (backbone)\\
\midrule
CIFAR-100~\cite{cifar10} &\textbf{82.67} &81.60~\cite{Bello_ICCV19} (Wide-ResNet-28-10)\\
Oxford-102~\cite{oxford102} &\textbf{97.80} &91.10~\cite{Chen_arXiv21} (ResNet152-SAM)\\
Oxford-37~\cite{oxford37} &\textbf{91.80} &91.60~\cite{Chen_arXiv21} (ResNet50-SAM)\\
\bottomrule
\end{tabular}
\vspace{-.5em}
\caption{The comparison of classification accuracy (\%) between ISKD (backbone: ResNet-18~\cite{resnet}) and the prior works using backbones with more parameters.}
\label{ISKD_result2}
\vspace{-.5em}
\end{table}

\section{Experimental result}
\label{sec:exp_result}

The experimental results are summarized in Table~\ref{ISKD_result}, where we refer to each column by the experiment ID $E_1\sim E_9$. All the numbers are reported in the format of [accuracy]$_{[e_s]}$, where $e_s$ is the number of epochs which the student model is trained for. $E_1\sim E_6$ list the performance of $S_1\sim S_6$, and $E_7$ summarizes the last performance after $i_1\sim i_6$. There is no teacher model for $E_1$ and the large-epoch baseline ($E_8$), but for the baseline using Tf-KD$_{self}$~\cite{Yuan_CVPR20} ($E_9$), the teacher model is the trained $S_1$. All the student models are pre-trained from ImageNet~\cite{imagenet} for ISKD ($E_7$) and the two baselines ($E_8$, $E_9$).

In Table~\ref{ISKD_result}, the accuracy of $E_2$ is higher than that of $E_1$, which shows that self KD can improve the student model's accuracy. The accuracy of $E_p$ is higher than that of $E_q$ for most cases when $2\leq q < p\leq 6$, which validates the assumption that in self KD, better teacher models result in better student models and that self KD can be done iteratively instead of just once. Comparing $E_7$ with $E_8$ and $E_9$, we show that given a fixed number of training epochs, ISKD outperforms the large-epoch baseline and the state-of-the-art self KD method Tf-KD$_{self}$~\cite{Yuan_CVPR20} in most cases. The fact that $E_7$ outperforms $E_9$ is also an ablation study supporting that self KD is better done iteratively than just once. 

Since Table~\ref{ISKD_result} covers diverse task domains, including pothole, generic, fine-grained, and medical imaging classification, our results support that ISKD can serve as a general-purpose performance booster. In addition, we obtain the results in Table~\ref{ISKD_result} by directly using the $\alpha$ values chosen for pothole classification \textit{without} tuning the $\alpha$ values for each dataset, which supports that the $\alpha$ values we use are transferable across different datasets. We also compare the accuracy of ISKD (backbone: ResNet-18~\cite{resnet}) with the prior methods which use the backbones with more parameters in Table~\ref{ISKD_result2}, where ISKD performs on par or even outperforms the listed methods which use more parameters. This finding supports that ISKD is more parameter efficient than the listed methods.

Furthermore, we analyze what is the worst performing teacher model in \textit{self} KD which can still make the student model outperform the baseline with no KD ($E_1$ in Table~\ref{ISKD_result}). To gain more insight, we design the following experiment to find the accuracy relation between the teacher and student models. Given that $E_1$ in Table~\ref{ISKD_result} is trained for 50 epochs, we use the 50 models saved after each epoch is completed as the teacher models. We repeat $E_2$ in Table~\ref{ISKD_result} for 50 times (each time with one of the 50 teacher models produced during $E_1$), and record the teacher-student accuracy relation. We perform this experiment on the simplex~\cite{simplex_dataset} and complex~\cite{complex_dataset} datasets using ResNet-18~\cite{resnet} as the network architectures.

\begin{figure}
\centering
\includegraphics[width=1\linewidth]{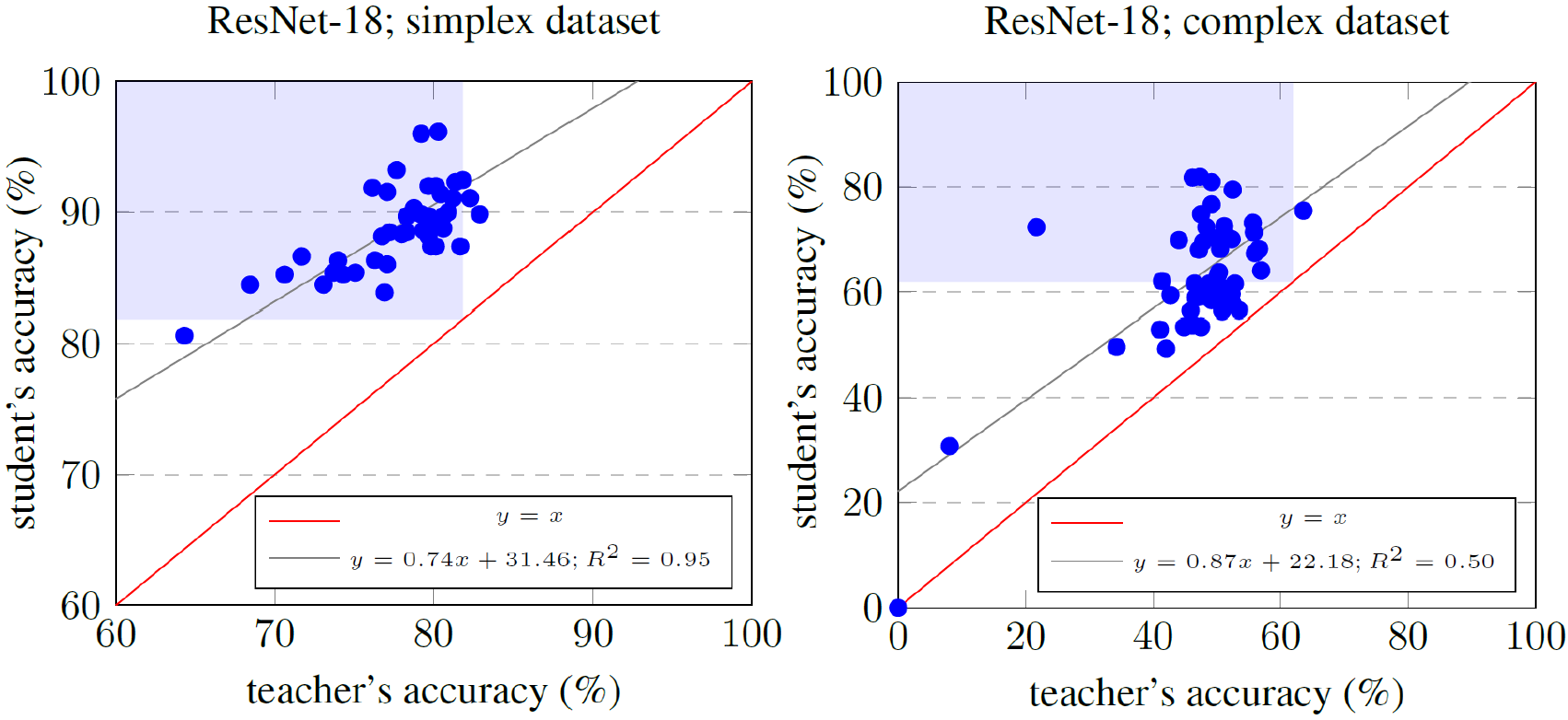}
\vspace{-1.5em}
\caption{The teacher-student accuracy relation on the simplex~\cite{simplex_dataset} and complex~\cite{complex_dataset} datasets using ResNet-18~\cite{resnet}. The gray lines are obtained from linear regression, and the line equation and Pearson's correlation coefficient $R$ are marked in the legend. Given the baseline performance in $E_1$ of Table~\ref{ISKD_result}, the shaded blue areas are the areas where the teacher model performs worse than the baseline but the student model outperforms the baseline.}
\label{ts_relation}
\vspace{-1em}
\end{figure}

We show the result of teacher-student accuracy relation in Fig.~\ref{ts_relation}. Most of the data points are above the red line ($x=y$), which supports that performing self KD can make the student model outperform the teacher model. Fig.~\ref{ts_relation} suggests that the accuracy of the teacher and student models has strong positive correlation (the slopes of the gray lines are positive and the $R^2\ge0.5$), which again supports the assumption that using better teacher models in KD generally results in better student models. We find that the number of data points falling into the shaded blue areas is not negligible, which serves as statistically more significant evidence than~\cite{Yuan_CVPR20} supporting that even if the teacher model is worse than the baseline, it is still likely that the student model can outperform the baseline after KD.

\begin{figure}	
\includegraphics[width=1\linewidth]{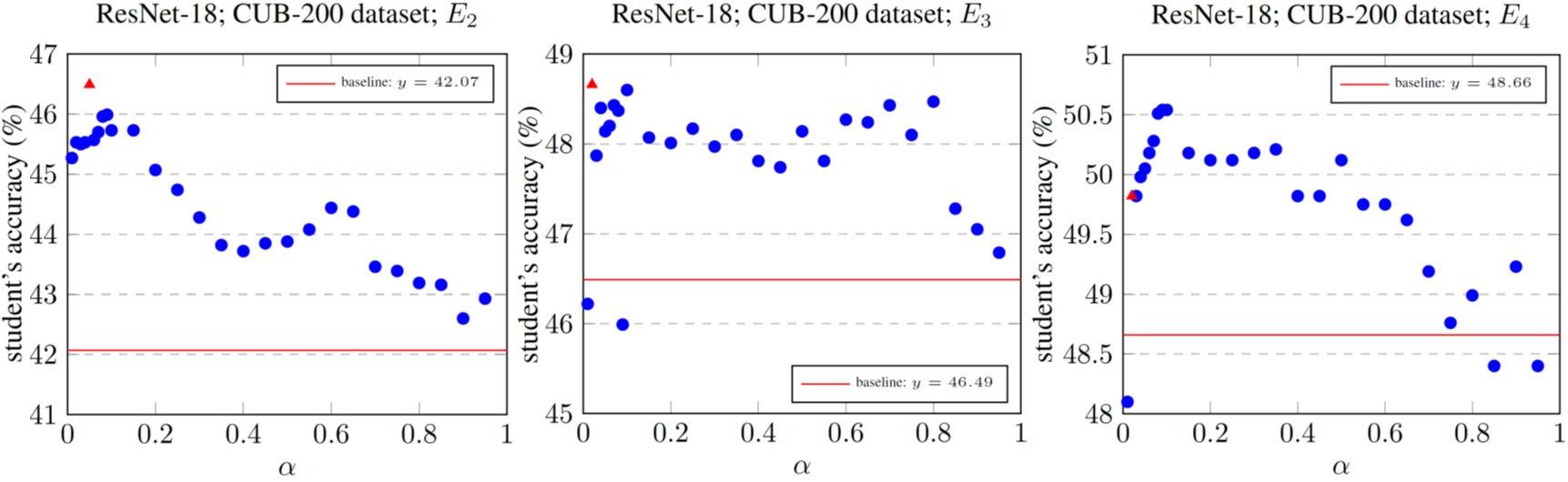}
\vspace{-1.5em}
\caption{The accuracy of the student model using ResNet-18 under different $\alpha$ values when we repeat $E_2$, $E_3$, and $E_4$ on the CUB-200 dataset. The triangular red points are the accuracy we report in Table~\ref{ISKD_result} by using the $\alpha$ values chosen for pothole classification. Shown as the red lines, the baseline for $E_i$ is the accuracy of the student model in $E_{i-1}$ reported in Table~\ref{ISKD_result}.}
\label{diff_alpha}
\vspace{-1em}
\end{figure}

Another experiment which is also not presented in the paper of Yuan~\etal~\cite{Yuan_CVPR20} is the impact of the $\alpha$ values on the accuracy. To study this, we use the ResNet-18~\cite{resnet} as the network architecture of ISKD and the same random seed 1, repeat $E_2$, $E_3$, and $E_4$ corresponding to the CUB-200~\cite{cub200} dataset with different $\alpha$ values, and report the student's accuracy in Fig.~\ref{diff_alpha}, where the red lines mark the baseline accuracy (\ie, the student's accuracy in the previous KD iteration reported in Table~\ref{ISKD_result}). For each sub-figure of Fig.~\ref{diff_alpha}, the teacher model is fixed as the corresponding one used in Table~\ref{ISKD_result}. The triangular points in Fig.~\ref{diff_alpha} are the accuracy reported in Table~\ref{ISKD_result} using the $\alpha$ values chosen for pothole classification, so their corresponding accuracy is not necessarily the best. Fig.~\ref{diff_alpha} shows that the student model outperforms the baseline in each KD iteration of ISKD under a wide range of $\alpha$ values, which supports that ISKD is flexible in terms of parameter selection.

\section{Conclusion}

We propose iterative self knowledge distillation (ISKD) in self KD to improve pothole classification accuracy when training lightweight models given a fixed amount of training epochs. Experimenting on three pothole classification datasets and six other datasets associated with generic classification, fine-grained classification, and medical imaging application, we show that ISKD outperforms the state-of-the-art self KD method Tf-KD$_{self}$, for most cases, given the same number of training epochs and that ISKD has wide applicability to various tasks without the need of a given teacher model and extra trainable parameters. In addition, we show more evidence supporting that the performance of the student model can benefit from self KD even when the pre-trained student model (which serves as the teacher model) is only moderately trained. Our study on the impact of the weight balancing the objectives of ISKD shows that even if we choose different weights deviating from the weights we initially use within a reasonable range, the student model can still improve over KD iterations, which supports that ISKD is flexible in terms of parameter selection.

\bibliographystyle{IEEEbib}

\end{document}